\documentclass{article}

\usepackage{times}
\usepackage{graphicx} 
\usepackage{subfigure} 
\usepackage{comment}
\usepackage{caption}

\usepackage{booktabs} 
\usepackage[space]{grffile}

\usepackage{amsmath}
\usepackage{amsthm}
\usepackage{amsfonts}

\usepackage{natbib}

\usepackage{algorithm}
\usepackage{algorithmic}
\usepackage{color}

\usepackage{hyperref}

\usepackage{dblfloatfix}    

\newcommand{\crossentropy}[2]{\mathrm{H}(#1 \| #2)}
\newcommand{\entropy}[1]{\mathrm{H}(#1)}

\usepackage[accepted]{icml2018}

\icmltitlerunning{Kickstarting Deep Reinforcement Learning}

\begin{document} 

\twocolumn[
\icmltitle{Kickstarting Deep Reinforcement Learning}
\icmlsetsymbol{equal}{*}

\begin{icmlauthorlist}
\icmlauthor{\hspace{2.1cm}Simon Schmitt}{equal,dm}
\icmlauthor{Jonathan J.~Hudson}{equal,dm}
\icmlauthor{Augustin Zidek}{equal,dm}\newline
\icmlauthor{Simon Osindero}{dm}
\icmlauthor{Carl Doersch}{dm}
\icmlauthor{Wojciech M. Czarnecki}{dm}
\icmlauthor{Joel Z. Leibo}{dm}
\icmlauthor{Heinrich Kuttler}{dm}
\icmlauthor{Andrew Zisserman}{dm}
\icmlauthor{Karen Simonyan}{dm}
\icmlauthor{S. M. Ali Eslami}{dm}
\end{icmlauthorlist}
\icmlaffiliation{dm}{DeepMind, London, UK}
\icmlcorrespondingauthor{Simon Schmitt}{suschmitt@google.com}

\icmlkeywords{distillation, reinforcement, learning, reinforcement learning, kickstarting, transfer learning}

\vskip 0.55in
]



\printAffiliationsAndNotice{\icmlEqualContribution} 

\begin{abstract}

We present a method for using previously-trained `teacher' agents to kickstart the training of a new `student' agent. To this end, we leverage ideas from policy distillation~\cite{rusu2015policy,parisotto2015actor} and population based training~\cite{jaderberg2017pbt}. Our method places no constraints on the architecture of the teacher or student agents, and it regulates itself to allow the students to surpass their teachers in performance. We show that, on a challenging and computationally-intensive multi-task benchmark~\cite{beattie2016dmlab}, kickstarted training improves the data efficiency of new agents, making it significantly easier to iterate on their design. We also show that the same kickstarting pipeline can allow a single student agent to leverage multiple `expert' teachers which specialise on individual tasks. In this setting kickstarting yields surprisingly large gains, with the kickstarted agent
matching the performance of an agent trained from scratch
in almost 10$\times$ fewer steps,
and surpassing its final performance by 42\%. Kickstarting is conceptually simple and can easily be incorporated into reinforcement learning experiments.

\end{abstract} 
\vspace*{-0.4cm}

\section{Introduction}

\begin{figure*}[t]
\vskip 0.2in
\begin{center}
\includegraphics[width=.99\textwidth]{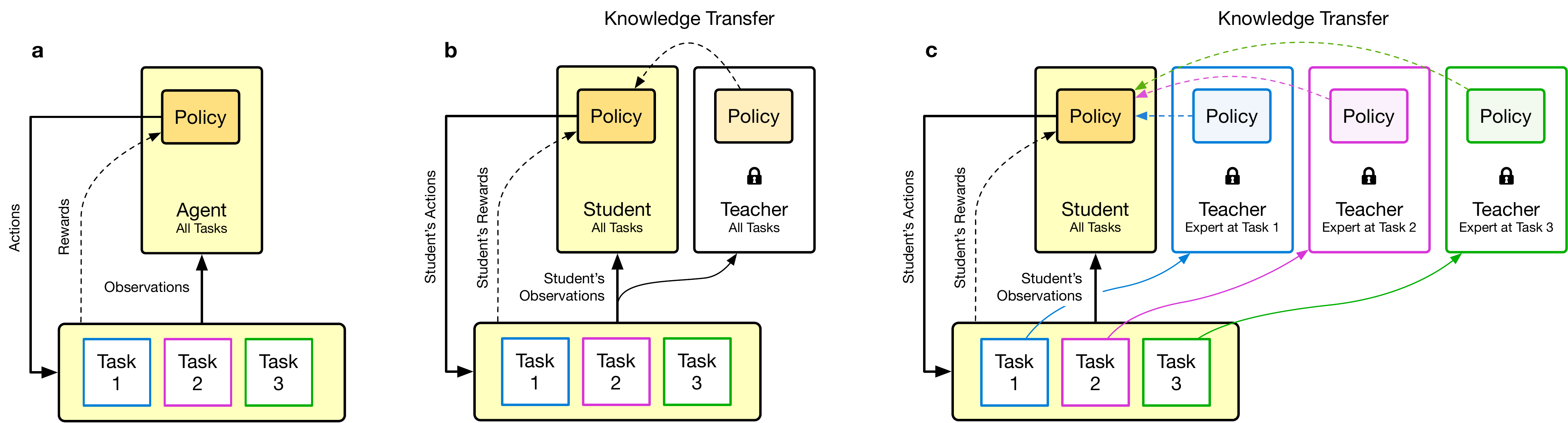}
\caption{\textbf{Overview.} Schematics of the three agent training scenarios explored in this paper. \textbf{a:} Standard multi-task RL. An agent learns to perform well on several tasks by acting on and observing all of them.  \textbf{b:} Single-teacher kickstarting (section \ref{experiments-single}). Observations from all tasks are also fed to a fixed, previously-trained teacher, and knowledge is transferred from teacher to student by encouraging the student to match the teacher's actions.  \textbf{c:} Multiple-teacher kickstarting (section \ref{experiments-multiple-experts}). For each task, observations are sent to a task-specific expert teacher; the student is encouraged to match this teacher's actions on the task.
}
\label{fig:overview}
\end{center}
\end{figure*}

Learning from teachers is a hallmark of human development, partly since gathering experience in the environment is time-consuming and bears significant risks~\cite{henrich2015secret}. In recent artificial intelligence research, ever-increasing task complexity is raising the cost of learning; with modern agents requiring up to billions of environment steps to achieve competitive performance from scratch \cite{impala}. Therefore, there is great interest in considering how agents might effectively learn from other agents, using them as teachers.

Despite the prevalence of weight transfer in supervised deep learning, such as pre-training networks on
ImageNet~\cite{girshick2014rich,Oquab2014LearningAT}, there has been limited success in leveraging previously trained agents in the reinforcement learning (RL) setting. In this paper we advocate a new approach, \textit{\mbox{kickstarting}}, where the explicit goal is to train new `student' agents rapidly when in the presence of previously trained `teacher' agents. In essence, the kickstarted student agent asks: ``What would my teacher do if it were in my shoes?'', and is encouraged to act in a similar manner. This differs from imitation learning \cite{argall2009survey} where an agent learns from the recorded experiences of an expert, not the expert itself.

We combine ideas from policy distillation \cite{rusu2015policy} and population based training \cite{jaderberg2017pbt}, yielding a method that is conceptually simple and can easily be incorporated into reinforcement learning experiments. Our method does not prescribe any architectural constraints on the student or teacher agents. Unlike in standard distillation, the framework automatically adjusts the influence of the teachers on the student agents, allowing the students to rapidly surpass their teachers in performance.
 
We present the following results, using the IMPALA agent~\cite{impala}, and experimental evaluation on the DMLab-30 task suite~\cite{beattie2016dmlab}:
(i)~We show that by kickstarting with a single teacher agent, we achieve up to 1.5$\times$ speedup over training a state-of-the-art agent from scratch on a challenging multi-task benchmark.
(ii)~We show that the student agent can rapidly outperform its teacher.
(iii)~We show that by kickstarting with multiple, task-specific expert teacher agents, we realise significant  gains, with the kickstarted agent matching the performance of an agent trained from scratch in 9.58$\times$ fewer steps, and surpassing its final performance by 42.2\%.
Taken together, our experiments demonstrate conclusively that kickstarting can significantly speed up the pace of research.

\section{Kickstarting RL Agents}

We consider the reinforcement learning setting where \emph{pre-trained} agents are readily available, which is often the case in practice (e.g.\ such agents could be previous versions of the agent solving the same task, or `expert' agents specialising in sub-tasks). We wish to train a new student agent with an arbitrary architecture in a way that both optimises the usual RL objective (i.e.\ expected return) and at the same time makes use of the pre-trained teacher agent in a way that leads to faster training and/or higher rewards.
To address these desiderata, we propose a framework called kickstarting, depicted in Figure~\ref{fig:overview}.

The main idea is to employ an auxiliary loss function which encourages the student policy to be close to the teacher policy on the trajectories sampled by the student. Importantly, the weight of this loss in the overall learning objective is allowed to change over time, so that the student can gradually focus more on maximising rewards it receives from the environment, potentially surpassing the teacher (which might indeed have an architecture with less learning capacity). 

In multi-task problems, it is also straightforward to extend this approach to the case of multiple teachers, each of which is an expert on a particular task: in this case the student will learn from an appropriate teacher on each task using an analogous formulation.

\subsection{Knowledge Transfer}

The core of kickstarting is a knowledge transfer mechanism which allows a student network to exploit access to (possibly multiple) expert teachers. One of the well known approaches to this problem is policy distillation~\cite{rusu2015policy} which defines it as a supervised learning problem.
Namely, a teacher policy $\pi_T$ is used to generate trajectories $x$,
each containing a sequence of states $(x_t)_{t \ge 0}$, over which one tries to match student's policy $\pi_S$, parameterised by $\omega$, to 
$\pi_T$. The corresponding loss function term for each sequence $s$ and each time step $t$ is:
\begin{equation}
\label{eq:distloss}
l_\mathrm{distill}(\omega, x, t) = \crossentropy{\pi_T(a|x_t)}{\pi_S(a|x_t, \omega)},
\end{equation}
where $\crossentropy{\cdot}{\cdot}$ is the cross-entropy.

Unlike in policy distillation, we do not wish for the student agent to merely replicate the behaviour of the teacher, but rather to maximise its own future rewards; teacher policies should only be proxies, helping to achieve this goal.

We begin with a standard RL objective~\cite{sutton-barto98} of maximising the expected return $\mathbb{E}_{\pi_S}\left[ R \right]$ over trajectories generated by the student agent; the return is defined as a sum of future discounted rewards: 
$R=\sum_{t\ge0} \gamma^t r_t$, where $\gamma\in[0,1)$ is the discount factor, $r_t$ is the reward received at time $t$, and $a_t\sim\pi_S$ is the action sampled from the student policy.
We consider policy-based model-free RL methods which optimise $\mathbb{E}_{\pi_S}\left[ R \right]$ by gradient ascent w.r.t.\ $\omega$ using some estimate of the policy gradient. 
This is usually represented as a loss $\ell_\mathrm{RL}(\omega, x, t)$ over a trajectory $x$ sampled from some behaviour policy.

Kickstarting adds to the above loss a term which is the cross-entropy between the teacher and student policies, weighted at  optimisation iteration $k$ by the scaling $\lambda_k \geq 0$:
\begin{equation}
\label{eq:kickloss}
l^k_\mathrm{kick}(\omega, x, t) = \ell_\mathrm{RL}(\omega, x, t) + \lambda_k  \crossentropy{\pi_T(a|x_t)}{\pi_S(a|x_t, \omega)},
\end{equation}
where $x$ is a trajectory generated by following the \emph{student} policy $\pi_S$.
Introduction of $\lambda_k$ allows an agent to initially focus on the supervision provided by the teacher. This ensures a dense learning signal, and does not have to be fully aligned with the RL objective. 
Then through adaptation of $\lambda_k$ during the course of learning, the agent is able to shift its optimization focus on the (potentially sparse) reward signal $r_t$,
similar to how continuation methods relax optimisation problems to make finding the solution easier~\cite{gulcehre2016mollifying, mobahi2016training}.

The kickstarting loss is closely related to policy distillation (as there is a knowledge transfer from teacher to student), but it differs in a few key aspects from~\eqref{eq:distloss} defined in~\cite{rusu2015policy}:
\begin{itemize}
    \item The student is solely responsible for generating trajectories, and can explore parts of the state space that the teacher does not visit. We note that student trajectories have also been used for distillation in~\cite{parisotto2015actor} which is otherwise similar to~\cite{rusu2015policy} and still does not incorporate the RL objective.
    \item The knowledge transfer process changes over time due to $\lambda_k$. In particular, after a certain number of iterations $T_0$ we would like to have $\lambda_k = 0 \; \forall k>T_0$ so that eventually the student becomes independent of the teacher and acts purely to maximise its own rewards.
    \item The resulting formulation still uses reinforcement learning and is not reduced to supervised learning. The student is no longer forced to exactly replicate the teacher's behaviour. As we will show in Section~\ref{seq:exp_results} the presence of the RL component in the objective leads to higher rewards compared to using the distillation loss only.
\end{itemize}

Our auxiliary loss can also be seen from the perspective of entropy regularisation. In the A3C actor-critic method~\cite{A3C2016} one adds the negated entropy
$\entropy{\pi_S(a|x_t, \omega)}$
as an auxiliary loss to encourage exploration.
But minimisation of negated entropy is equivalent to minimising the KL divergence $\mathrm{D}_\mathrm{KL}(\pi_S(a|x_t, \omega) \| U)$, where $U$ is a uniform distribution over actions. Similarly the kickstarter loss is equivalent to the KL divergence between the teacher and the student policies.
In this sense, the kickstarter loss can be seen as encouraging behaviour similar to the teacher, but just as entropy regularisation is not supposed to lead to convergence to a uniform policy, the goal of kickstarting is not to converge to the teacher's policy. The aim of both is to provide a helpful auxiliary loss, based on what is a sensible behaviour -- for the case of entropy regularization it is just sampling a random action, while for kickstarting it is following the teacher.

In the following sections we elaborate on the terms comprising the full learning objective and the corresponding update rules (Section~\ref{sec:kick_ac}), and describe how one can adjust $\lambda_k$ in an automatic, reward-driven manner (Section~\ref{sec:pbt}).

\subsection{Kickstarting Actor-Critic}
\label{sec:kick_ac}

We begin with a brief recap of the actor-critic RL objective~\cite{A3C2016}.
The updates for the policy can be expressed as a simple loss over the sampled trajectory and consists of two terms: the first corresponds to reward maximisation (i.e. the policy gradient term) and the second corresponds to entropy regularisation (a term to encourage exploration). 
Using the notation introduced in previous sections it can be written as:
\begin{equation}
\begin{aligned}
\ell_\mathrm{A3C}(\omega, x, t) = &\log \pi_S(a_t|x_t,\omega)(r_t + \gamma v_{t+1} - V(x_t| \theta)) \nonumber \\ 
&- \beta \entropy{\pi_S(a|x_t, \omega)}
\end{aligned}
\end{equation}
where $x$ are sampled from the learner's policy (i.e. the trajectories are on-policy), $\beta$ is the exploration weight, $v_{t+1}$ is the value function target, and $V(x_t;\theta)$ is a value approximation computed by the critic network parameterised by $\theta$ ($\theta$ and $\omega$ do not have to be disjoint, e.g.\ when the policy and critic are implemented as separate heads on top of a shared network, which is the case here).
The critic is updated by minimising the loss to the value target $\|V(x_t;\theta)-v_t\|_2^2$.

Consequently, the A3C kickstarting loss becomes:
\begin{equation}
\ell_\mathrm{A3C}(\omega, x, t) + \lambda_k  \crossentropy{\pi_T(a|x_t)}{\pi_S(a|x_t, \omega)}.
\end{equation}
Note that the $\ell_\mathrm{A3C}$ term contains an entropy cost, which encourages the student to explore beyond the teacher's supervision.

IMPALA~\cite{impala} extends the actor-critic formulation of~\cite{A3C2016} to a large-scale setting with distributed worker machines.
Multiple actor workers interact with copies of the environment, taking actions according to the current policy, and generating trajectories of experience.
These trajectories are sent to a learner worker, which computes the policy update on a batch of trajectories.
Due to the lag between sampling actions (by the actors) and computing the update (by the learner), IMPALA's actor-critic algorithm (called V-trace) is off-policy, and consequently needs to employ a correction (based on importance sampling). 

V-trace introduces importance sampling weights $\rho_t$ to the policy gradient for a student trajectory, $x$, corresponding to maximising the expected return at step $t$:
\begin{equation}
\rho_t \nabla_\omega\log\pi_S(a_t|x_t,\omega) \big( r_t+\gamma v_{t+1} - V(x_t|\theta)\big).
\end{equation}
Apart from this modification, it follows the actor critic loss and thus incorporation of Kickstarting into IMPALA is analogous to any other actor-critic system. The only difference is that the trajectories used for the kickstarting loss now come from the \emph{actors} of the student policy, rather than directly from the student.

\subsection{Population Based Training}
\label{sec:pbt}

An important element of the proposed method is the adjustment of the schedule of kickstarter loss weights ($\lambda_k$ in Eq.~\ref{eq:kickloss}). While this can be done manually -- similarly to how learning rate schedules are often hand-crafted for a given application -- it requires additional expert knowledge of the problem, and a method of trial and error, which is time-consuming. This problem becomes even more pronounced if we have $K$ teachers, rather than a single teacher; thereby requiring tuning for $K$ different schedules.

An alternative, which we employ, is to use an online hyperparameter tuning technique: the recently proposed Population Based Training~\cite{jaderberg2017pbt} (PBT). 
This method trains a population of agents in parallel in order to jointly optimise network weights and hyperparameters which affect learning dynamics (such as the learning rate or entropy cost). 
In our instantiation of PBT, each agent periodically selects another member of the population at random and checks whether its performance is significantly better than its own. If this is true, weights and hyperparameters of the better agent are adopted. Independently of the outcome, with fixed probability, each hyperparameter is slightly modified for the purpose of exploration.

While using PBT is not essential to get the benefits of kickstarting, we found it useful in practice, as it helps to further decrease the time it takes to perform research iterations. 
In particular, it allows us to automatically adjust the schedule for $\lambda_k$ separately for each teacher in a multi-teacher scenario, while simultaneously adjusting the learning rate as well as the entropy regularisation strength. 

\section{Related Work}

The idea of having experts which can be used to train new agents through matching the output distributions was explored originally for supervised learning and model compression~\cite{hinton2015distilling, BaDistill14, bucilua2006model}, and was subsequently adapted for multitask reinforcement learning. Typically one gathers experience from expert policies, which are then used to train a student model using supervised learning~\cite{rusu2015policy, berseth2018}. Consequently the focus has hitherto been on compression and teacher-matching, rather than the ultimate goal of reward maximisation. 

Similarly in the Actor-Mimic~\cite{parisotto2015actor} approach, one tries to replicate the teacher's behaviour to later exploit it for transfer learning. In departure from \cite{rusu2015policy}, the Actor-Mimic collects experience from the student policy, and this is empirically shown to be beneficial for replication of the teacher's behaviour. Although it is not explored in these papers, after performing distillation one could fine-tune the student policy using rewards. 

From this perspective kickstarting can be seen as a continuous version of such two-phase learning, with a focus on reward maximisation from the very beginning (which does not require arbitrary stopping criteria for any of the phases, as it is a joint optimisation problem).

In this sense it is even more related to the dynamics of learning emerging from  Distral~\cite{teh2017distral}. However there the focus is on the performance of teachers, and a student is used purely as a communication channel between them (as one jointly tries to distill knowledge from teachers to the student, and the student regularises the teachers so they do not move too far away from its policy). 
Finally, the kickstarting approach
has similarities to the `never-ending learning'
paradigm~\cite{mitchell2015never,chen2013neil,carlson2010toward}, which has long argued that
machines should gradually accumulate knowledge from diverse
experience and over long timeframes.

\section{Experimental Setup}

All experiments were based on the IMPALA agent~\cite{impala}, trained on the DMLab-30 task suite~\cite{beattie2016dmlab}. We first give an outline of the construction of the agent. We follow that with a short description of the task suite.

\subsection{IMPALA}

We experiment with the same two agent network architectures as in~\cite{impala}. 
Both architectures have a convolutional pathway for visual features, followed by an LSTM, the ouptut of which is fed to policy and value heads. These heads are simple linear layers.  
For the tasks where language input is available, there is an additional language LSTM whose output is concatenated with the output of the visual network. The `small' agent is intended as a simpler baseline, which is fast to train, and has 2 convolutional layers. The `large' agent represents the state of the art for these tasks; it has 15 convolutional layers, and significantly outperforms the small agent.

\subsection{Population Based Training}

In our experiments each population member comprised one high-capacity learner worker (running on a Nvidia P100 GPU) and 150 actor workers, with each actor running a single copy of the environment. The workers were distributed equally, 5 per task, between the 30 tasks of the suite. 

For the multi-teacher setup, we use a separate $\lambda^{i}_{k}$ for each teacher, and allow PBT to choose a schedule for each.  
However, we expect that the distillation weights will be correlated.  Hence we implement the weights in factorised way: $\lambda^{i}_{k} = \alpha_k \rho^{i}_{k}$, so that the evolution algorithm can make all weights stronger or weaker simultaneously.

\subsection{DMLab-30 Suite}

We evaluate our agents on the DMLab-30 task suite~\cite{beattie2016dmlab}. 
DMLab-30 consists of 30 distinct tasks that are designed to challenge a broad range of agent ``cognitive capabilities''. For example, the suite includes tasks involving navigation through procedurally-generated mazes (requiring short- and long-term memory of maze structure, and in some cases planning and prediction), tasks of foraging for food items in naturalistic environments (challenging the visual system of an agent), and tasks based on classical experiments from cognitive neuroscience like visual search~\cite{leibo2018psychlab}. 
Due to the breadth of tasks and the complexity of the 3D environments, this suite comprises a challenging and comprehensive benchmark for agent development.

We measure the performance of each agent on DMLab-30 by calculating the mean capped human-normalised score. We record the average return for all episodes in a window of $30$ million environment frames that the IMPALA learner experiences. We normalise the per task score relative to human performance, where human performance is considered to be 100. These scores are then capped at 100. This capping provides us with a metric that helps focus on producing agents that have human level competency across the entire suite, rather than allowing, for instance, an agent to merely exploit its super-human reaction times and consequently achieve a very high score due to significant skew on a small subset of tasks.
To obtain a score for the full suite, we average this human-normalised score across all 30 tasks.

\section{Experimental Results}
\label{seq:exp_results}

We present agent kickstarting results using a single teacher and using multiple teachers separately, in section \ref{experiments-single} and \ref{experiments-multiple-experts} respectively.

\subsection{Kickstarting With a Single Teacher}
\label{experiments-single}

\begin{figure}
    \centering
    \includegraphics[width=.95\columnwidth]{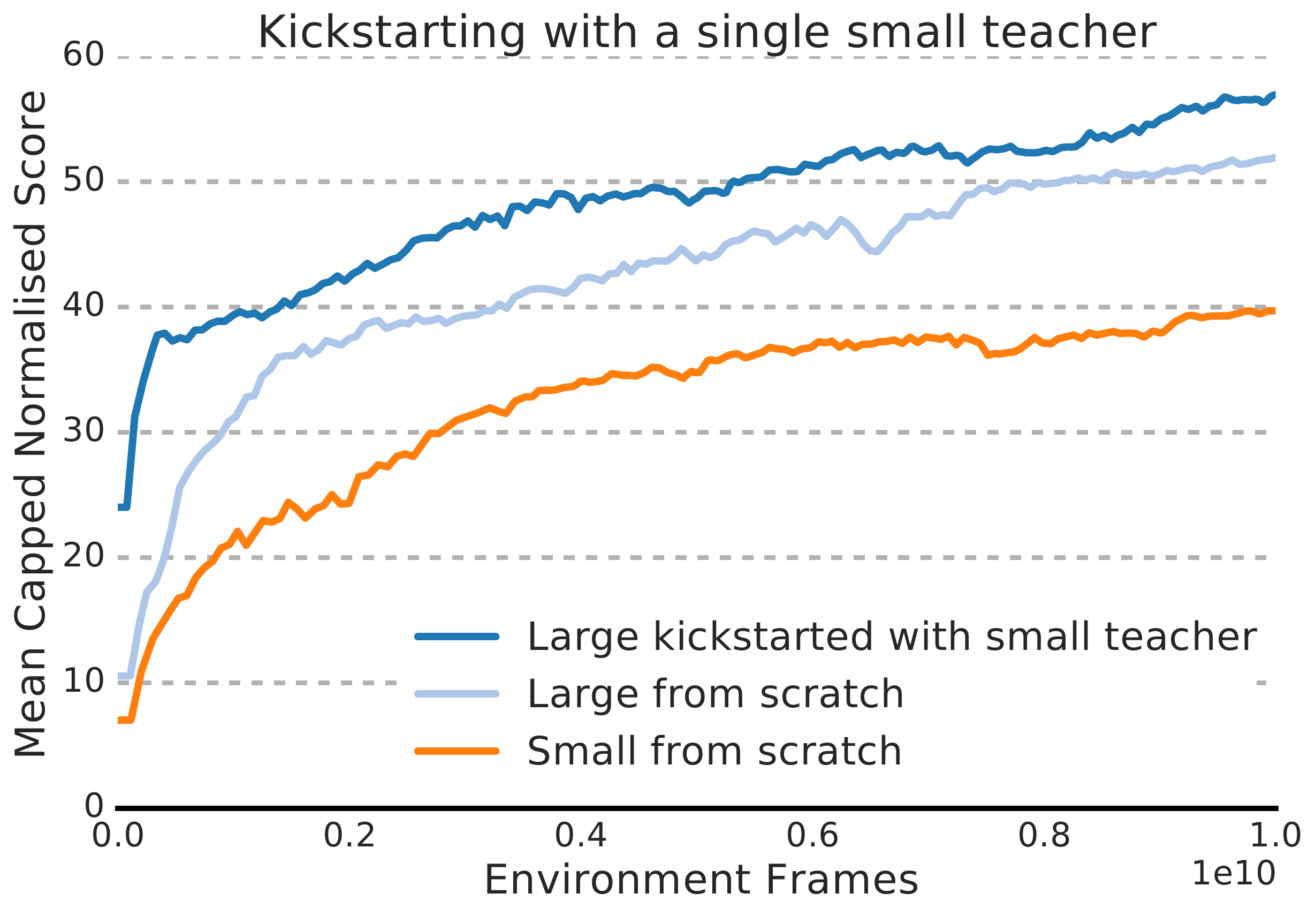}
    \includegraphics[width=.95\columnwidth]{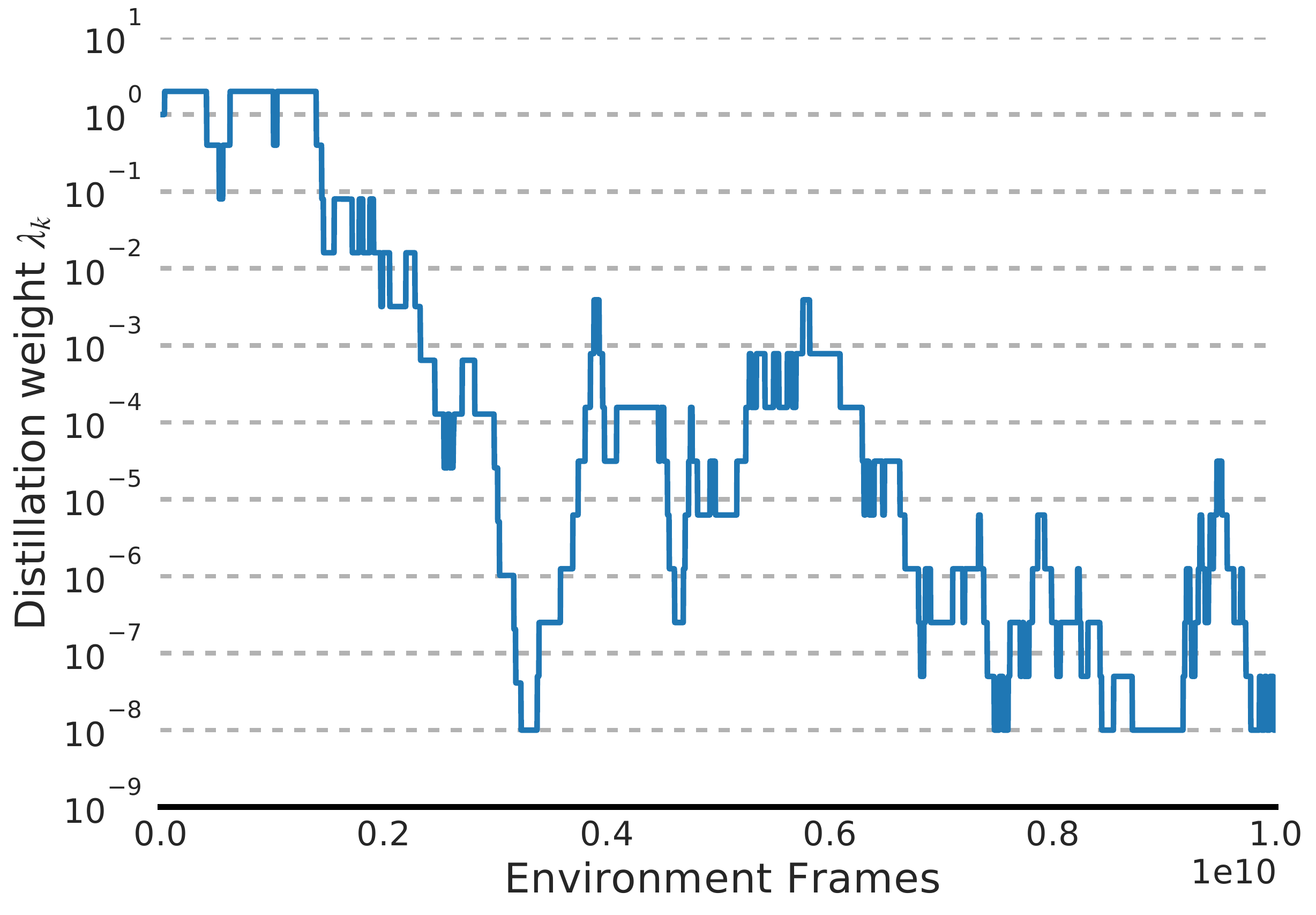}
    \caption{Performance of agents on all DMLab-30 tasks. We observe that the kickstarted agent learns faster than small and large from-scratch agents. \textbf{Top:} Rolling average of agent's score (mean of top-3 PBT population members each). \textbf{Bottom:} PBT evolution of kickstarting distillation weight over the course of training for the best population member. See also Table~\ref{kickstarting_single_shallow_teacher_table}.}
    \label{kickstarting_single_shallow_teacher}
\end{figure}

\begin{table*}
    \caption{Benefits of kickstarting with a single small teacher.  We report score after a fixed number of frames have been experienced, and the total number of frames required in order to reach fixed levels of performance. 
    See also Figure~\ref{kickstarting_single_shallow_teacher}.
    }
    
 \centering
    \begin{tabular}{l|llll|lll}
    \toprule
     & \multicolumn{4}{c|}{\textbf{Score at Frames}} &  \multicolumn{3}{c}{\textbf{Frames to Reach Score}} \\
     &      0.5B &      1.0B &      2.0B &      10.0B &	    30.0 &            40.0 &            50.0 \\
    \midrule
    Large kickstarted &             37.4 &             39.4 &             42.4 &             56.9  &      0.13B &\
     1.39B &     5.31B \\
    Large from scratch  &             24.1 &             31.1 &             37.5 &             51.9 &    0.99B & \
    3.26B &     8.14B \\
    \midrule
    \textbf{Improvement}                               &  \textbf{+55.2\%} &  \textbf{+26.7\%} &  \textbf{+13.1\%}\
 &  \textbf{+9.6\%} & \textbf{6.92}\small $\times$ &   \textbf{2.35}\small $\times$ &  \
 \textbf{1.53}\small $\times$ \\
    \bottomrule
    \end{tabular}

\label{kickstarting_single_shallow_teacher_table}
\end{table*}

We experiment with kickstarting from a single small teacher to a large student agent. This resembles the typical architecture exploration use-case where a large architecture sweep takes place after a small preliminary study.  Figure~\ref{kickstarting_single_shallow_teacher}  compares kickstarted training to from-scratch training performance over time.  We observe a substantial speedup: $6.92\times$ to reach a score of $30$, and $1.53\times$ to reach a score of $50$.
In fact, it takes the large student only about 1 billion frames to reach the small teacher's final (after 10 billion frames) performance.  Table~\ref{kickstarting_single_shallow_teacher_table} gives a speedup break-down.

Figure~\ref{kickstarting_single_shallow_teacher} also shows the schedule for the kickstarting distillation weight over the same time period, showing a reduction to almost negligible weight after 2 billion steps. 
At that point the student is essentially free, and it continues to learn and surpasses the teacher by a margin of $+43.4\%$ ($56.9$ vs.\ $39.7$ DMLab-30 score).

\begin{figure}
    \centering
    \includegraphics[width=.95\columnwidth]{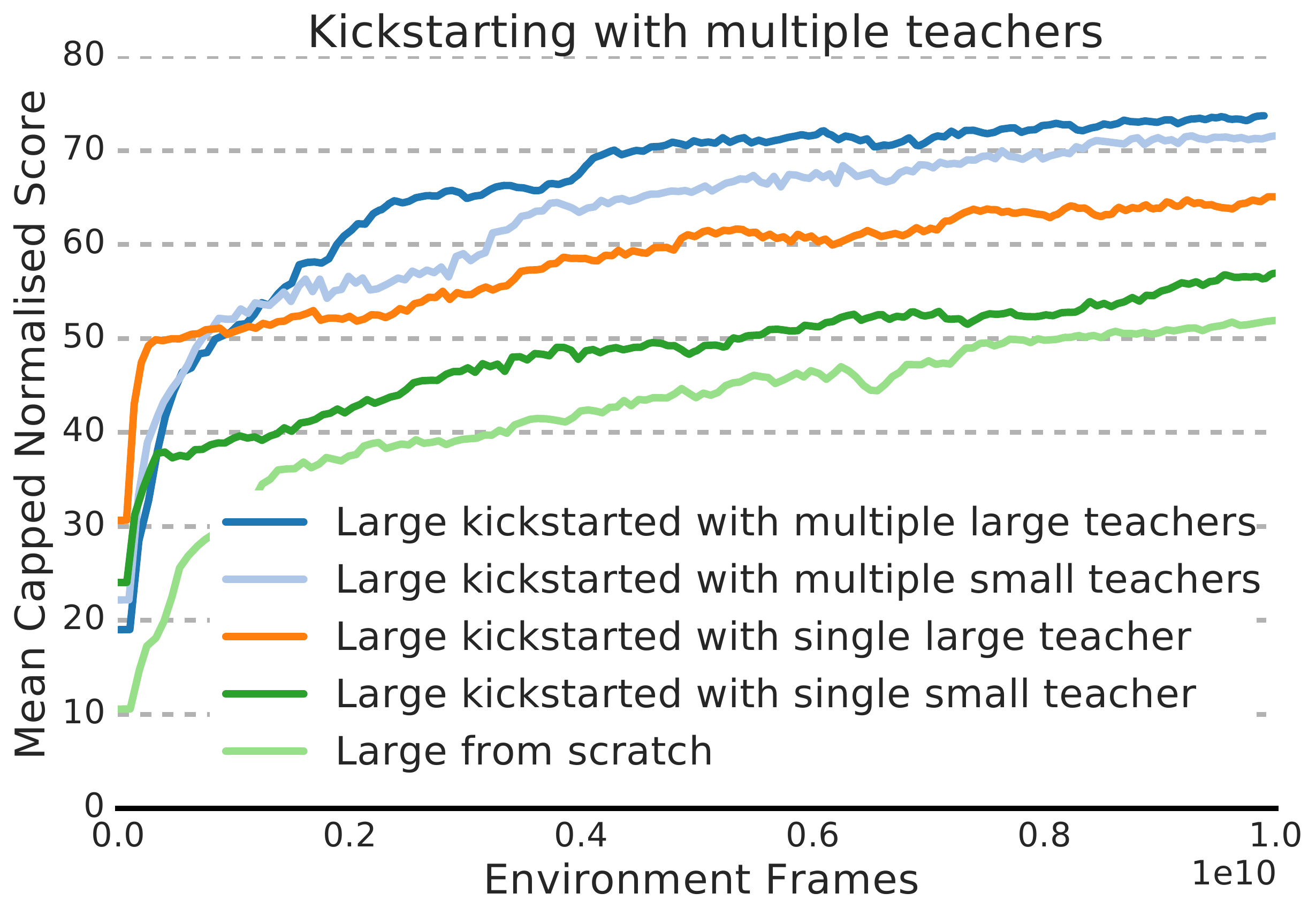}
    \caption{
    Comparison of performance for agents kickstarted with a single (large or small) teacher and with multiple (large or small) teachers, showing a significant boost for both multiple teachers and for larger teachers.  Kickstarting always improves over multi-task training from scratch. 
    The reported metric is the average DMLab-30 score among the top-3 PBT members. 
    }
    \label{kickstarting_multiple_teacher}
\end{figure}

We also experimented with kickstarting a large student agent with a large teacher trained from scratch. Figure~\ref{kickstarting_multiple_teacher} shows that this agent achieves higher performance than the equivalent agent kickstarted by a small teacher. This is as expected, reflecting the better tuition from the large teacher agent, as compared to the small teacher.

\subsubsection{PBT and Other Distillation Weighting Approaches}
\label{experimental_single_teacher_pbt}

\begin{figure}
    \centering
    \includegraphics[width=.95\columnwidth]{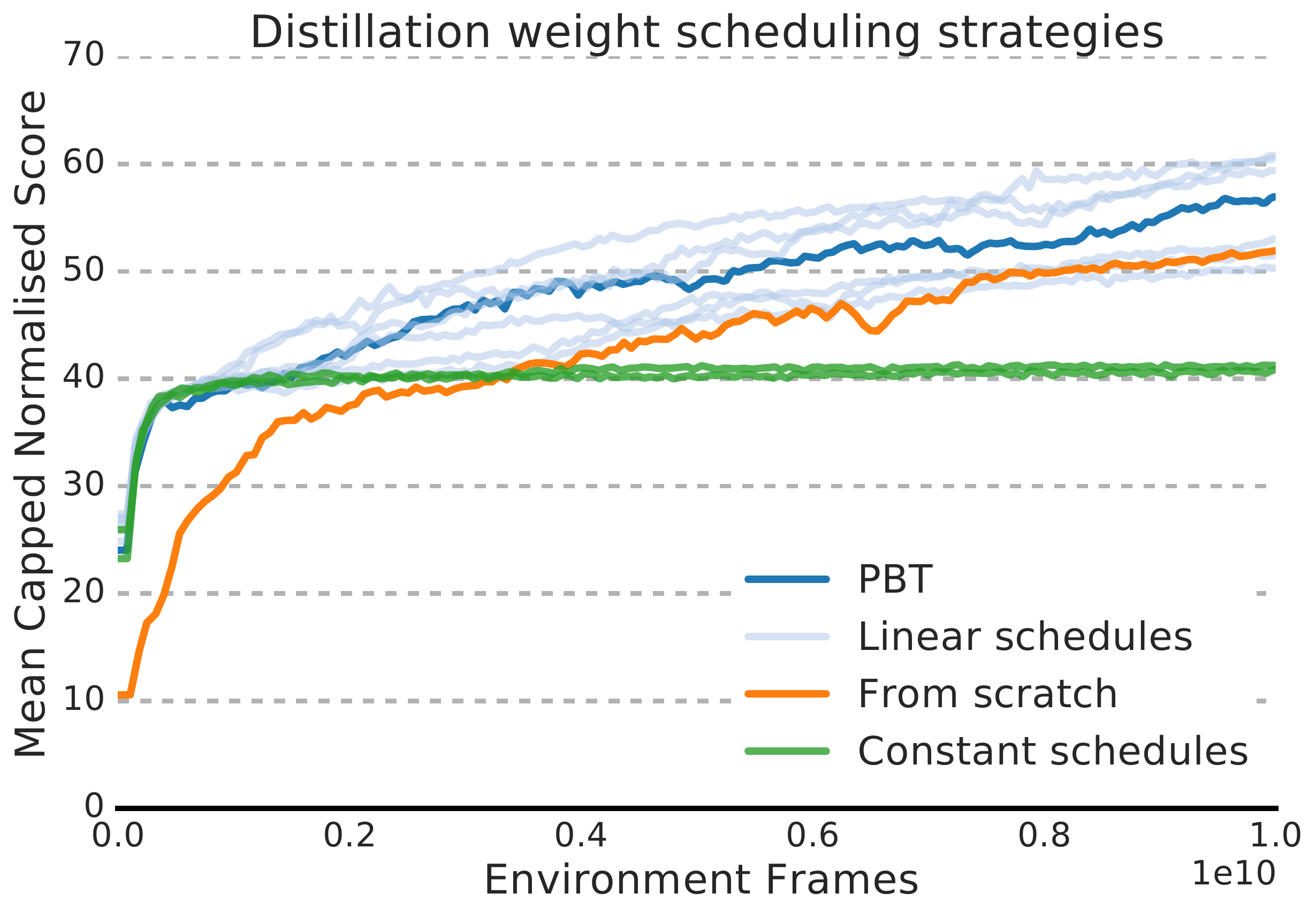}
    \caption{
    Performance of different strategies for changing the distillation weight $\lambda_k$ over the course of training. 
    Agents with a constant setting of $\lambda_k$ perform well at first, but their performance ultimately plateaus. Linear schedules can work well, but only if chosen correctly. PBT performance is competitive and alleviates the need for sweeps over such schedules. The reported metric is the average DMLab-30 score of the top-3 population members for each method. The `From Scratch' curve shows the performance of the teacher agent for comparison. See also Table~\ref{kickstarting_single_shallow_teacher_schedule_table}.}
    \label{kickstarting_single_shallow_teacher_schedule_figure}
\end{figure}

\begin{table}[t]
    \caption{
    Final performance for different distillation weight scheduling strategies.
    The reported metric is average DMLab-30 score of the top-3 population members at $10$B frames. See also Figure~\ref{kickstarting_single_shallow_teacher_schedule_figure}.}
    \centering
    \begin{tabular}{lll}
    \toprule
    \textbf{Score at Frames} & 1.0B & 10.0B \\
    \midrule
    \textbf{Baselines} \\
    \midrule
    Small from scratch (teacher)  & 21.6   &   39.7 \\ 
    Large from scratch & 31.1  &   51.9 \\ 
    \midrule
    \textbf{Kickstarting: PBT} \\
    \midrule
    Large kickstarted with small teacher & 39.4 &   \textbf{56.9} \\ 
    \midrule
    \textbf{Kickstarting: Constant Schedules} \\
    \midrule
    Schedule: constant = 1            & 40.0     &   41.2 \\ 
    Schedule: constant = 2            & 39.6      &   40.8 \\  
    \midrule
    \textbf{Kickstarting: Linear Schedules} \\
    \midrule
    Schedule: linear from 1 to 0 at 1B   & 41.3  &   \textbf{59.4} \\ 
    Schedule: linear from 1 to 0 at 2B    & 40.1   &   \textbf{60.8} \\ 
    Schedule: linear from 1 to 0 at 4B     & 39.9  &   53.1 \\ 
    Schedule: linear from 2 to 0 at 1B    & 40.7   &   \textbf{60.5} \\ 
    Schedule: linear from 2 to 0 at 2B    & 40.0  &   51.5 \\ 
    Schedule: linear from 2 to 0 at 4B    & 39.2 &   50.3 \\ 
    \bottomrule
    \end{tabular}
    \label{kickstarting_single_shallow_teacher_schedule_table}
\end{table}

\begin{table*}
    \centering
    
    \caption{\
    Benefits of kickstarting with multiple teachers, analogous to Table~\ref{kickstarting_single_shallow_teacher_table}.  We report score after a fixed number of frames have been experienced, and the total number of frames required to reach fixed levels of performance.
    See also Figure~\ref{kickstarting_multiple_teacher}.}
    \begin{tabular}{l|llll|lll}
    \toprule
     & \multicolumn{4}{c|}{\textbf{Score at Frames}} &  \multicolumn{3}{c}{\textbf{Frames to Reach Score}} \\
     &      0.5B &      1.0B &      2.0B &      10.0B &	    30.0 &            50.0 &            70.0 \\
    \midrule
    Large kickstarted &             44.8 &             51.0 &             61.3 &             73.8  &      0.21B &\
     0.85B &     4.27B \\
    Large from scratch  &             24.1 &             31.1 &             37.5 &             51.9 &    0.99B & \
    8.14B &     - \\
    \midrule
    \textbf{Improvement}                               &  \textbf{+85.9\%} &  \textbf{+64.0\%} &  \textbf{+63.5\%}\
 &  \textbf{+42.2\%} &  \textbf{4.29}\small $\times$ &  \textbf{9.58}\small $\times$ &  \
 \textbf{-}\small \\
    \bottomrule
    \end{tabular}
    
    \label{kickstarting_multiple_teacher_table}
\end{table*}

As mentioned, we expect the distillation weight $\lambda_k$ to be important, especially as the student's performance approaches the teacher's.
Thus, we consider ways to control $\lambda_k$, including various hand-crafted schedules and constants.
Results  are shown in Table \ref{kickstarting_single_shallow_teacher_schedule_table} and Figure \ref{kickstarting_single_shallow_teacher_schedule_figure}. 
Note that even for the manually-specified schedules, PBT still controls other hyperparameters (learning rate and entropy cost), as described in \cite{impala}.
In general, constant schedules can perform well early on, but ultimately cause the agent to plateau near the teacher's final performance, suggesting that the agent is trying too hard to match the teacher.
Linear schedules that reduce $\lambda_k$ to 0 tend to work better, although it is important to reduce this weight quickly, a fact that was not apparent {\it a priori}.
Using PBT to control this hyperparameter works nearly as well as the best manually-specified schedules, suggesting that it is an efficient alternative to running a sweep across schedules.

\subsection{Kickstarting With Multiple Teachers}
\label{experiments-multiple-experts}

We next extend kickstarting to a scenario where we have multiple `expert' teachers, potentially trained separately on different tasks.  
We expect that the ensemble of experts will together have considerably more knowledge about the task suite, and so ideally the student will absorb even more knowledge than it could from a single teacher.

Figure \ref{kickstarting_multiple_teacher} and Table \ref{kickstarting_multiple_teacher_table} show our overall results in this scenario.  
Indeed, the kickstarted agent achieves high performance, far higher than an agent trained from scratch, and even higher than an agent kickstarted with a single teacher (73.8 vs 56.9 for a comparable single teacher setup).
This suggests that kickstarting works well even when combining the expertise of many diverse agents, which means that multi-expert kickstarting is a good way to learn a single agent that performs well on a large set of tasks.

\subsubsection{Expert Kickstarting vs. Distillation}

\begin{figure}[htb]
    \centering
    \includegraphics[width=0.95\columnwidth]{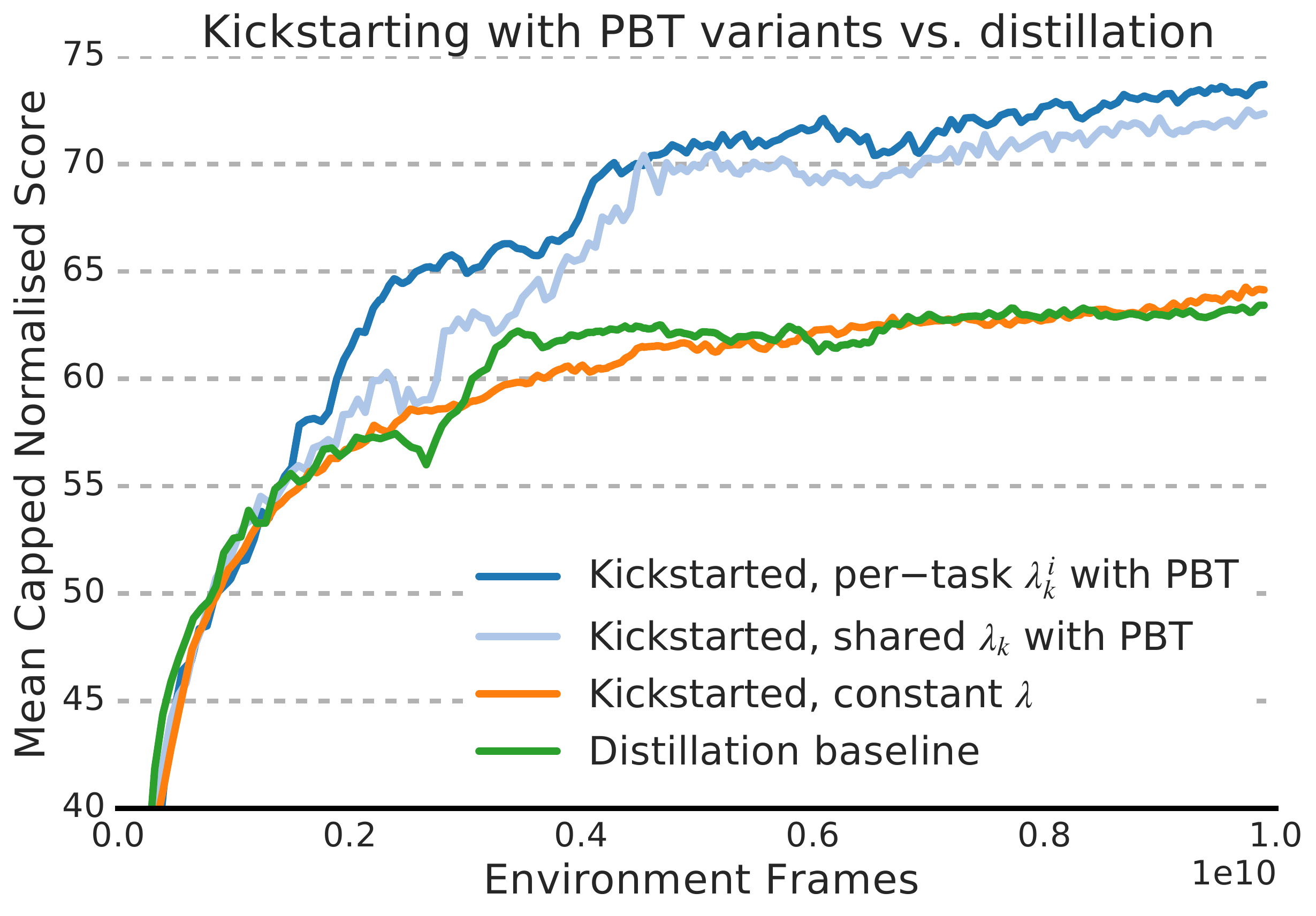}
    \caption{
    Comparison of four approaches for knowledge transfer from multiple large teachers: kickstarting with three PBT parametrizations for the distillation weight $\lambda$ (constant weight across all tasks $\lambda$, evolved shared weight for all tasks $\lambda_{k}$, evolved per-task weight $\lambda^{i}_{k}$), and a pure policy distillation agent without any reinforcement learning loss.}
    \label{kickstarting_multiple_deep_teacher_options}
\end{figure}

Figure~\ref{kickstarting_multiple_deep_teacher_options} shows a comparison between agents kickstarted with multiple teachers and an agent of the same architecture trained purely with 
the distillation loss~\eqref{eq:kickloss}, without any of the typical reinforcement learning losses, and thus similar to Policy Distillation and Actor-Mimic~\cite{rusu2015policy,parisotto2015actor}.
Kickstarting outperforms distillation by a significant margin, with the gap widening as training time increases.
This suggests that it is useful for the student to learn from its own rewards as well as the teacher's policy, and indeed necessary for the student to outperform the teacher.

\subsubsection{Analysis of Distillation Weight $\lambda_k$}

Thus far we have considered a single approach to setting $\lambda_{k}$: a separate $\lambda^{i}_{k}$ for the $i$'th teacher, where PBT automatically selects a schedule for each.  
This follows from the intuition that the student may surpass different experts at different times. However, one might ask how important it is to tune schedules separately in practice.
Figure~\ref{kickstarting_multiple_deep_teacher_options} explores alternatives where a shared $\lambda_{k}$ is used for all teachers and updated via PBT, or where a constant $\lambda$ is used for all teachers.  
While a fixed schedule still performs poorly, we find that a separate $\lambda^{i}_{k}$ for each teacher provides a negligible boost, suggesting that the procedure is not exceptionally sensitive to this hyperparameter.

\begin{figure*}[htb]
    \centering
    \includegraphics[width=1.942\columnwidth]{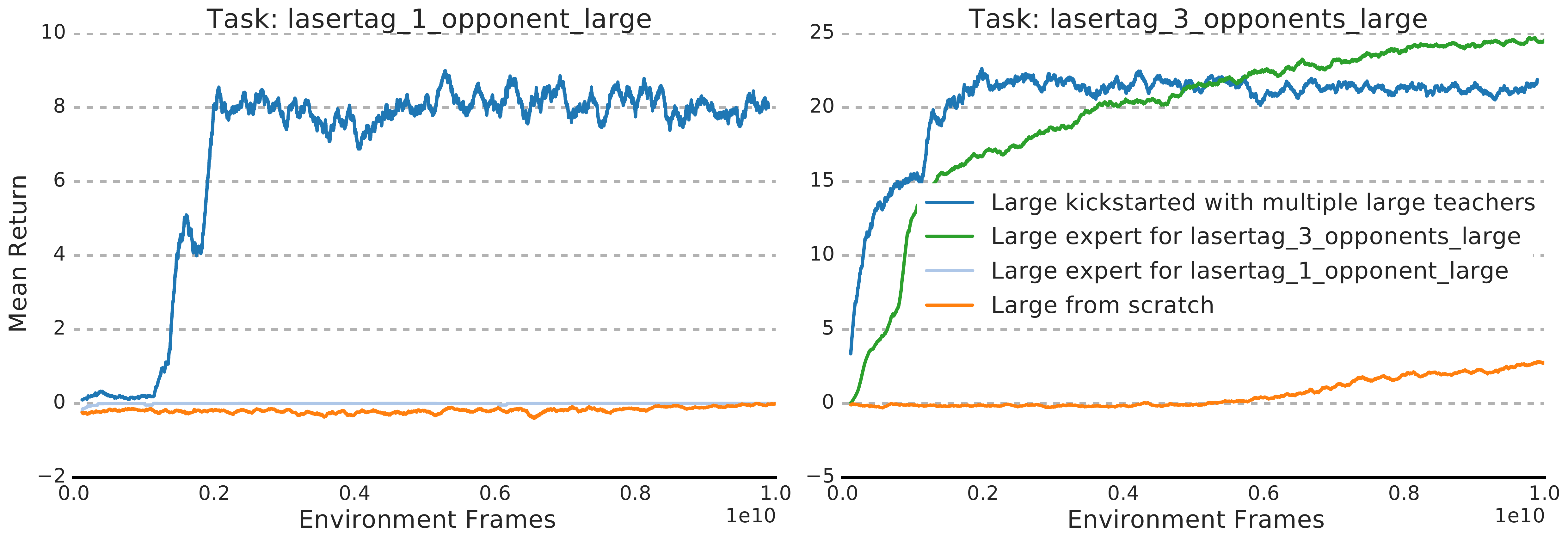}
    \caption{
    Performance of a from-scratch multi-task agent, two single-task expert agents (distinct for each task), and a multi-task agent kickstarted with experts on two laser-tag tasks.
    The multitask from-scratch agent learns neither task.
    The expert on the (easier) 3-bot task learns, whereas the agent trained solely on the (harder) 1-bot variant does not. 
    The student kickstarted using the experts masters both the 1- and 3-bot variants, indicating transfer from the 3- to the 1-bot environment.}
    \label{fig:lasertag}
    \vspace{0.7cm}
\end{figure*}

\subsubsection{Per Task Performance}

We briefly analyze agent performance on a few DMLab-30 tasks to gain insight into our performance gains when kickstarting with multiple teachers. 
Two kinds of tasks, laser tag and navigation, are particularly informative.

\begin{figure}[H]
    \centering
    \includegraphics[width=.971\columnwidth]{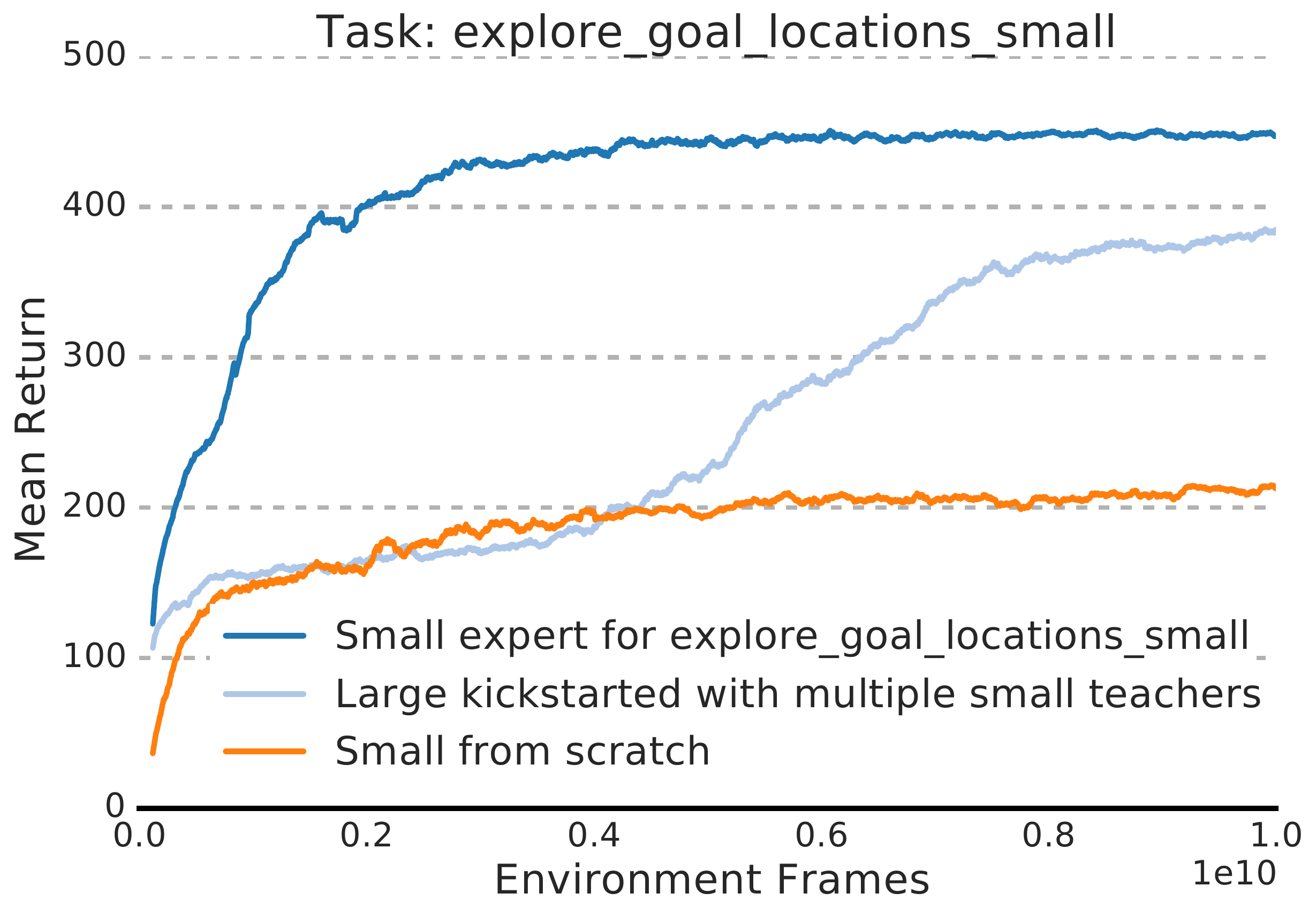}
    \caption{
    Performance of a from-scratch multi-task agent, a single-task expert, and a multi-task agent kickstarted with experts on a navigation task.
    The multitask from-scratch agent fails to acquire short-term memory, while the agent kickstarted with expert teachers does thanks to the expert. }
    \label{fig:navigation}
\end{figure}

The task suite contains a set of similar `laser tag' tasks in which agents must navigate a procedurally-generated maze and tag other opponent `bots' with a laser. 
Figure~\ref{fig:lasertag} 
shows the performance of three types of agent:  a multi-task agent kickstarted with expert teachers; a multi-task agent trained from-scratch; and a single-task expert agent.
In the 1-bot variant, encounters with the single opponent (and thus rewards) are very sparse: thus the from-scratch agent and even the single-task expert do not learn.
The multi-teacher kickstarted agent, however, learns quickly.
The reason is that a single-task expert learned strong performance on the 3-bot task (thanks to denser rewards), and its knowledge transfers to the 1-bot variant.
In fact, we find the student `ignores' the (incompentent) 1-bot expert (its $\lambda^{i}_{k}$ quickly goes to 0).
Figure~\ref{fig:navigation} shows the performance on the ``explore goal locations small'' navigation task. 
Here, the agent must repeatedly navigate a random maze: for several trials, the spawn position is random, but the goal location and maze layout remain the same.
Agents that cannot remember the maze layout after respawning--such as the from-scratch agent in Figure~\ref{fig:navigation}--cannot score higher than 300 on this task.

\vspace{-1cm}
The expert learns to use its short term memory and scores considerably higher and, thanks to this expert, the kickstarted agent also masters the task. 
This is perhaps surprising because the kickstarting mechanism only guides the student agent in which action to take: it puts no constraint on how the student structures its internal memory state.
However, the student can only predict the teacher's behaviour by remembering information from before the respawn, which seems to be enough supervisory signal to drive short-term memory formation.
We find this a wonderful parallel with how the best human educators teach: not telling the student what to think, but simply putting the student in a fruitful position to learn for themselves.

\section{Conclusion}

We have presented kickstarting -- a training paradigm that helps both shorten the cycle-time for research iterations in deep RL, and that helps student agents achieve performances that exceed those attained by agents trained from scratch. The method is simple to implement in policy-based reinforcement learning setups.
In contrast to policy distillation, our method allows students to actively balance their own learning objectives with the advice given by the teachers. This feature allows the students to surpass the performance of their teachers. 

More fundamentally, we believe this result opens the pathway to a new research direction, where new agents are designed to be adept in absorbing and using the knowledge of previously trained agents. Indeed, it might be possible for this scheme to lead to the training of complex agents that would not have been possible to train from scratch in the absence of their particular ancestral lineage.

\section*{Acknowledgements}

We thank Lasse Espeholt, Hubert Soyer, and Chloe Hillier for helpful discussions, advice and support.

\vspace{0.5cm}

\bibliography{example_paper}
\bibliographystyle{icml2018}

\end{document}